\documentclass[11pt]{article}
\usepackage{amsfonts}
\usepackage[preprint]{acl}
\usepackage{amsmath}
\usepackage{pifont}
\usepackage{times}
\usepackage{latexsym}
\usepackage{url}
\usepackage{tcolorbox}
\tcbuselibrary{breakable}
\usepackage{enumitem} % Required for the formatted bullet points
\usepackage{multicol}
\usepackage{lmodern}
\usepackage[T1]{fontenc}
\usepackage{listings}
\usepackage{multirow}
\usepackage[T1]{fontenc}
\usepackage[utf8]{inputenc}

\usepackage{microtype}
\usepackage{booktabs}  
\usepackage{inconsolata}
\usepackage[ruled,vlined]{algorithm2e}

\usepackage{graphicx}
\newcommand{\cmark}{\ding{51}}% ✓
\newcommand{\xmark}{\ding{55}}% ✗
\title{\texttt{AMEND++}: Benchmarking Eligibility Criteria Amendments in Clinical Trials}

\author{
  \textbf{Trisha Das\textsuperscript{1}},
  \textbf{Mandis Beigi\textsuperscript{2}},
  \textbf{Jacob Aptekar\textsuperscript{2}},
  \textbf{Jimeng Sun\textsuperscript{1}},
\\
  \textsuperscript{1}University of Illinois Urbana-Champaign,
  \textsuperscript{2}Medidata Solutions,
\\
  \small{
    \textbf{Correspondence:} \href{mailto:trishad2@illinois.edu}{trishad2@illinois.edu}
  }
}

\begin{document}
\maketitle
\begin{abstract}
Clinical trial amendments frequently introduce delays, increased costs, and administrative burden, with eligibility criteria being the most commonly amended component. We introduce \textit{eligibility criteria amendment prediction}, a novel NLP task that aims to forecast whether the eligibility criteria of an initial trial protocol will undergo future amendments. To support this task, we release \texttt{AMEND++}, a benchmark suite comprising two datasets: \texttt{AMEND}, which captures eligibility-criteria version histories and amendment labels from public clinical trials, and \texttt{AMEND\_LLM}, a refined subset curated using an LLM-based denoising pipeline to isolate substantive changes. We further propose \textit{Change-Aware Masked Language Modeling} (CAMLM), a revision-aware pretraining strategy that leverages historical edits to learn amendment-sensitive representations. Experiments across diverse baselines show that CAMLM consistently improves amendment prediction, enabling more robust and cost-effective clinical trial design.  Our code is publicly available at \url{https://anonymous.4open.science/r/amend-E61A/}.
\end{abstract}

\section{Introduction}

Clinical trial protocols define the scientific and operational blueprint of a study and must be followed exactly as approved by regulatory authorities. In practice, however, many trials undergo post-approval modifications (i.e., protocol amendments) to address emerging scientific questions, operational challenges, or regulatory feedback. While often necessary, amendments introduce substantial burdens by delaying timelines, requiring extensive coordination across investigators and sites, and generating high additional costs for sponsors and regulatory bodies.

A prior study from the Tufts Center for the Study of Drug Development has shown that amendments are both frequent and increasingly complex, with a growing proportion triggered by design deficiencies or changes in eligibility criteria \cite{ getz2016impact}. More recent analyses indicate a rise in amendment prevalence and longer implementation cycles \cite{getz2024new}, reflecting the intensification of regulatory expectations and operational demands. Regulatory agencies face similar pressures, processing large volumes of amendments each year, often with multi-week review periods. These findings underscore that many amendments could be anticipated earlier if potential issues in the initial protocol were detected proactively.

\begin{table*}[!t]
\centering
\small
\caption{Summary of existing studies on clinical trial protocol amendments and dataset availability. None of the prior studies released a public dataset suitable for machine learning tasks.}
\begin{tabular}{@{}p{3.2cm}p{3.5cm}p{2.5cm}p{4.3cm}p{1.2cm}@{}}
\toprule
\textbf{Study} & \textbf{Data Source} & \textbf{\# of Trials} & \textbf{Focus / Contribution} & \textbf{Public Dataset} \\ 
\midrule
\citep{getz2016impact} & 15 pharmaceutical companies and CROs (Tufts CSDD collaboration) & 836 & Quantified prevalence, causes, and costs of protocol amendments across commercial trials. & \xmark \\[4pt]

\citep{getz2024new} & 16 pharmaceutical companies and CROs & 950 & Updated benchmark analysis on amendment frequency and impact on study performance. & \xmark \\[4pt]

\citep{botto2024new} & \citep{getz2024new} & 950 & Compared amendment prevalence and completion outcomes in oncology vs.\ non-oncology trials. & \xmark \\[4pt]

\citep{joshi2023common} & 53 NHS-sponsored trials (UK) & 53 & Mixed-methods study identifying common amendment reasons and avoidability patterns. & \xmark \\[4pt]

\midrule
\texttt{AMEND} (ours) & \url{https://clinicaltrials.gov} & 161970 & Prepared an ML-ready benchmark dataset containing all versions of eligibility criteria and amendment labels. & \cmark \\[4pt]

\texttt{AMEND\_LLM} (ours)& \url{https://clinicaltrials.gov} & 64641 & Subset of \texttt{AMEND} dataset with denoised amendment labels using LLMs. & \cmark \\[4pt]
\bottomrule
\end{tabular}
\label{tab:amendment_summary}
\end{table*}

Eligibility criteria (EC)\footnote{Throughout this paper, we use EC to refer to the eligibility criteria section as a single textual unit.} are a particularly common driver of protocol amendments \cite{getz2024new}. Changes to inclusion and exclusion criteria can arise for multiple reasons, as noted in ICH E9 and prior clinical research \cite{Lewis1999StatisticalPFA, cleophas2009statistics}. New medical insights may necessitate amendments, while persistent violations of entry criteria or inadequate recruitment can also prompt revisions. Because modifications to EC directly affect who can enroll, changes introduced mid-trial can yield meaningful differences between participant populations before and after an amendment, with implications for study integrity and interpretability. These factors make EC amendments both impactful and important to anticipate early in the design process.

Despite the central role of protocol design in clinical trial efficiency, the computational literature lacks large-scale resources and models to predict whether EC in an initial protocol will be modified later (Table \ref{tab:amendment_summary}). EC define the study population and are among the most frequently amended components of trial protocols, making the initial EC text a strong signal for predicting future eligibility-criteria amendments. Automatically identifying EC at elevated risk of future amendments could help sponsors improve protocol quality, reduce avoidable revisions, and streamline trial execution.

\begin{comment}
To support research on clinical trial design, we introduce \texttt{AMEND++}, a benchmark suite to predict EC amendments from the initial protocol text. \texttt{AMEND++} comprises two complementary datasets: \texttt{AMEND}, a large-scale collection of clinical trials with eligibility amendment labels derived from publicly available protocol version histories, and \texttt{AMEND\_LLM}, a refined subset with high-quality labels that focus on substantive eligibility changes. Using the datasets, we benchmark several transformer-based models and show that the language of initial eligibility criteria contains predictive signals of future amendment risk.    
\end{comment}
To support research on clinical trial design, we introduce \texttt{AMEND++}, a benchmark suite for studying eligibility-criteria (EC) amendment prediction from initial trial protocols. \texttt{AMEND++} includes two datasets capturing different trade-offs between scale and label fidelity: \texttt{AMEND}, a large-scale collection derived from public protocol version histories, and \texttt{AMEND\_LLM}, a refined subset emphasizing substantive eligibility changes. Using this benchmark, we show that the information of initial eligibility criteria contains predictive signals of future amendment risk.

We make three primary contributions:
\begin{enumerate}
    \item We formulate \emph{eligibility criteria amendment prediction} as a new NLP task grounded in real-world clinical trial design, where the objective is to forecast whether an initial protocol will later undergo eligibility-related amendments.
    
    \item We release \texttt{AMEND++}, a benchmark suite consisting of two large-scale datasets: \texttt{AMEND}, which provides eligibility criteria version histories with amendment labels derived from public trial records, and \texttt{AMEND\_LLM}, a refined subset constructed using an LLM-based label denoising algorithm. This algorithm decomposes EC amendments into added, removed, and modified components to isolate substantive eligibility changes, resulting in high-quality amendment labels with substantially higher agreement with human annotations than alternative labeling approaches.

    \item We propose \emph{Change-Aware Masked Language Modeling} (CAMLM), a revision-aware pretraining strategy that leverages historical eligibility-criteria edits to learn amendment-sensitive representations. Through extensive benchmarks and ablations, we show that CAMLM consistently improves amendment prediction across multiple datasets, backbone encoders, and downstream classifiers.
\end{enumerate}

\begin{figure*}[!t]
    \centering
    \includegraphics[width=0.99\textwidth]{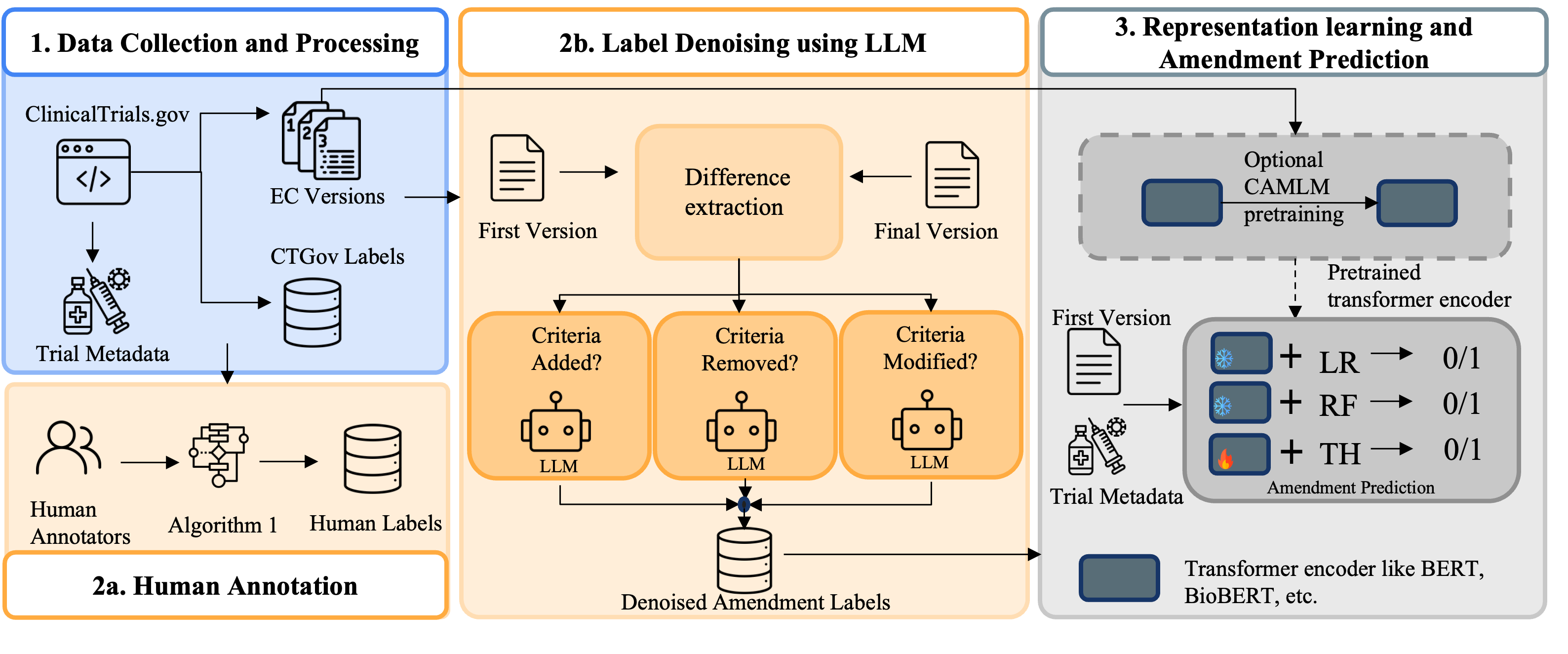}
    \caption{Overview of the proposed framework for clinical trial eligibility criteria (EC) amendment prediction, from data collection to downstream task application. LR: logistic regression, RF: random forest, TH: trial head used for predicting amendment. Human annotation is performed only on the test data.}
    \label{fig:pipeline}
\end{figure*}

\section{Related Work}
In recent years, a variety of machine learning and deep learning approaches have been applied throughout the clinical‐trial lifecycle, including outcome prediction \cite{gao2025auto, fu2022hint}, generation of digital twins or simulated trial participants \cite{das2023twin, das2025seqtrial}, and trial‐document similarity search \cite{das-etal-2025-secret, luo2024clinical}. These studies demonstrate the growing potential of data-driven methods to support trial design, analysis, and operational decision-making.

Few NLP systems have focused on extracting and normalizing eligibility criteria into structured representations, enabling cohort querying and downstream analysis but not modeling protocol evolution or amendment risk \cite{tseo2020information, yuan2019criteria2query}. More recently, multi-task benchmarks such as TrialBench have released standardized datasets spanning trial design and prediction tasks, including eligibility-criteria design, but do not consider revision-aware pretraining or forecasting protocol amendments from initial drafts \cite{chen2024trialbench}.

In contrast, research on protocol amendments has been primarily descriptive and statistical. Prior work has characterized the frequency, causes, and operational burden of amendments \cite{getz2016impact, botto2024new, joshi2023common}, but none have framed amendment forecasting as a predictive modeling problem or released ML-ready datasets. To our knowledge, no machine learning study has examined whether the text of initial protocols can predict future amendments. A closely related line of research examines the downstream statistical consequences of EC amendments. One study proposes conducting separate hypothesis tests for participants enrolled before and after an EC amendment and combining them using Fisher’s combination test, highlighting that population shifts induced by amendments can meaningfully affect inference \cite{losch2008statistical}. While this work focuses on post hoc analysis rather than prediction, it reinforces the broader importance of EC amendments and the need for methodological tools that account for changes in study populations.

\begin{comment}
 Taken together, existing literature underscores the operational and statistical impact of protocol amendments but provides no computational framework for predicting them. Our work addresses this gap by introducing large-scale datasets and benchmark models for forecasting amendment risk directly from eligibility-criteria text.   
\end{comment}

\section{Methods}
Our pipeline (Figure \ref{fig:pipeline}) builds the \texttt{AMEND++} benchmark from longitudinal eligibility-criteria histories, combines human annotation and LLM-based denoising for label construction, formulates amendment prediction as a supervised task, and introduces CAMLM for revision-aware pretraining.

\begin{comment}
\subsection{Data Collection}

We constructed a benchmark suite of interventional drug trials, which we refer to as \texttt{AMEND++}. All trials were obtained from \textit{ClinicalTrials.gov}. We first filtered for studies labeled as interventional and involving pharmacological interventions. The resulting corpus forms the basis of the \texttt{AMEND++} suite, which links each trial’s initial eligibility criteria to subsequent amendment outcomes. 

To create a higher-quality variant focused on substantive changes, we further processed trial version histories using large language models to identify and exclude purely administrative or minor edits. This yields \texttt{AMEND\_LLM}, a refined subset within \texttt{AMEND++} designed to support more precise modeling of meaningful eligibility‐criteria amendments.

Because historical versions of trial eligibility criteria (EC) are not available for download, we programmatically scraped the \textit{Record History} tab of each trial using \texttt{BeautifulSoup}. This allowed us to extract all historical versions of eligibility criteria for each trial. 

For each trial, we also collected additional protocol information that is typically available at the beginning of the trial, including trial title and description, disease indications, investigational drugs, MeSH terms corresponding to the diseases, phase, and other relevant metadata. This information provides features for downstream prediction tasks and ensures that each trial record contains a comprehensive set of information.   
\end{comment}

\subsection{Data Collection and Processing}

We constructed \texttt{AMEND++}, a benchmark suite of interventional drug trials curated from \textit{ClinicalTrials.gov}. We first filtered for pharmacological interventional studies and programmatically scraped the \textit{Record History} tab using \texttt{BeautifulSoup} to reconstruct longitudinal EC version histories, which are otherwise unavailable via standard API or direct downloads. 

The resulting suite comprises two datasets: \texttt{AMEND}, a large-scale collection capturing raw EC version histories and amendment labels directly from public records; and \texttt{AMEND\_LLM}, a refined subset where labels are denoised via an LLM-based pipeline to isolate substantive changes from minor administrative edits (e.g., formatting or typos). For each trial, we also extracted baseline metadata (e.g., titles, descriptions, disease indications, investigational drugs, MeSH terms, and study phases, etc.) to provide a comprehensive feature set for amendment prediction.

After scraping, we saved each version separately, preserving the full content and order of the original ECs. Disease names were standardized using MeSH terms to allow consistent comparisons across trials. For label denoising (see Section \ref{sec:label_denoising}), we only used the first and final EC versions. However, all versions are released for each clinical trial with EC amendments.

\subsection{Task Definition: Eligibility Criteria Amendment Prediction}
\label{subsec:task_definition}

We formulate \emph{eligibility criteria amendment prediction} as a supervised binary classification task. 
For each trial $i$, the model takes as input the initial eligibility criteria text $E_i^{(0)}$ and trial-level metadata $M_i$ available at trial initiation, and predicts $
\hat{y}_i \in \{0,1\}$,
where $\hat{y}_i = 1$ indicates that the eligibility criteria are amended in later protocol versions and $\hat{y}_i = 0$ otherwise.

\subsection{Label Denoising}
\label{sec:label_denoising}

\subsubsection*{Human Annotation}
To generate high-quality labels for EC amendments, we relied on human annotation only for the test split. For these trials, human annotators examined the first and final versions of the ECs and applied a set of systematic rules (Algorithm~\ref{alg:human_amendment}) to determine whether a substantive amendment had occurred. This split serves as a gold-standard evaluation set.

\subsubsection*{Label Denoising using LLM}
For the training and validation splits, large-scale manual annotation is prohibitively costly. Instead, we generate amendment labels using large language models (LLMs), resulting in the \texttt{AMEND\_LLM} dataset, a refined subset of the \texttt{AMEND} dataset.

We decompose amendment detection into three complementary, criterion-level checks, each handled by a separate LLM instance:

\begin{enumerate}
    \item \textbf{Criteria Added}: identifying whether any new inclusion or exclusion criteria are introduced in the final version.
    \item \textbf{Criteria Removed}: detecting whether any criteria present in the initial version are absent from the final version.
    \item \textbf{Criteria Modified}: identifying modifications to existing criteria that alter the eligible population, such as changes in numeric thresholds, medical entities, logical operators, or temporal constraints.
\end{enumerate}

This decomposition allows each LLM to focus on a single, well-defined type of change, reducing ambiguity and improving robustness compared to a single, monolithic amendment judgment.

For each check, we use a 1-shot prompting strategy, where the LLM is provided with a single illustrative example followed by the target input. To reduce context length and focus the LLMs on semantically meaningful changes, we compute textual differences between the first and final EC versions and provide only the extracted differences as input to the LLM. Instructions provided to the LLM also abide by the rules in Algorithm \ref{alg:human_amendment}. The prompts for LLMs are available in the Appendix (see prompts  \ref{fig: criteria_removed}, \ref{fig: criteria_modified}, \ref{fig: criteria_added}).

Formally, for a given trial $i$, let $E_i^{(0)}$ and $E_i^{(T)}$ denote the initial and final EC texts, respectively. We compute the differences:
\begin{equation}
    \Delta E_i = \textsc{Diff}(E_i^{(0)}, E_i^{(T)}),
\end{equation}
which serves as the LLM input for all amendment checks. Each LLM check produces a binary output:
\begin{equation}
    y_i^{(k)} = f_k(\Delta E_i), \quad k \in \{\text{add}, \text{remove}, \text{modify}\},
\end{equation}
where $f_k(\cdot)$ denotes the LLM corresponding to the $k$-th amendment type.

The final amendment label is obtained by aggregating the outputs of the three checks using a logical OR:
\begin{equation}
    y_i = \mathbb{I}\left( \bigvee_{k} y_i^{(k)} = 1 \right),
\end{equation}
where $\mathbb{I}(\cdot)$ is the indicator function.

Based on validation against human-annotated test data, the LLM-generated labels show strong agreement with human judgments and are therefore used for the training and validation splits of \texttt{AMEND\_LLM}. This strategy enables scalable and consistent denoising of amendment labels while reserving human annotation exclusively for evaluation.

\begin{table*}[!t]
\centering
\caption{Dataset statistics for \texttt{AMEND} and \texttt{AMEND\_LLM}.}
\label{tab:dataset_stats}
\resizebox{\textwidth}{!}{
\begin{tabular}{lcc}
\hline
\textbf{Property} & \texttt{AMEND} & \texttt{AMEND\_LLM} \\
\hline
\multicolumn{3}{l}{\textit{Train/Validation/Test Sizes}} \\
Train trials (1 / 0) 
& 140,312 (52,071 / 88,241) 
& 49,678 (17,178 / 32,500) \\

Validation trials (1 / 0) 
& 15,591 (5,846 / 9,745) 
& 8,896 (3,240 / 5,656) \\

Test trials (1 / 0) 
& 6,067 (2,786 / 3,281) 
& 6,067 (2,786 / 3,281) (same as \texttt{AMEND}) \\
\hline

\multicolumn{3}{l}{\textit{Label Sources}} \\
Train + validation labels 
& CTGov labels 
& LLM-denoised amendment labels \\

Test labels 
& Human annotations 
& Human annotations \\
\hline

\multicolumn{3}{l}{\textit{Amendment Count Statistics (amended trials only)}} \\
Min / Avg / Median / Max 
& 1 / 2.92 / 2 / 38 
& 1 / 3.07 / 2 / 38 \\
\hline
\end{tabular}}
\end{table*}

\subsection{Change-Aware Masked Language Modeling (CAMLM)}
\label{sec:camlm}

Standard masked language modeling (MLM) treats all tokens in a document as equally informative during pretraining. In contrast, EC amendments in clinical trial protocols are typically sparse and localized: only specific criteria are added, removed, or modified across versions. To exploit this structure, we propose a pretraining strategy, \emph{Change-Aware Masked Language Modeling} (CAMLM), that biases representation learning toward historically unstable regions of EC text. CAMLM does not introduce a new model architecture; instead, it modifies the masking policy used during MLM pretraining by leveraging historical versions of EC. The objective is to encourage the encoder to learn representations that are sensitive to content that has been revised in prior protocols, which may signal fragility or ambiguity in the initial trial design.

For each clinical trial $i$ with recorded EC version history, we construct ordered version pairs $\left(E_i^{(t)}, E_i^{(T)}\right)$, where $E_i^{(t)}$ denotes an earlier EC version ($t < T$) and $E_i^{(T)}$ denotes the final EC version. If multiple historical versions are available, each earlier version is paired with the final version, yielding multiple training instances that capture long-horizon protocol evolution. For trials without intermediate versions, we construct a single pair $\left(E_i^{(0)}, E_i^{(T)}\right)$, allowing both amended and non-amended trials to contribute to pretraining.

Given a version pair $\left(E_i^{(t)}, E_i^{(T)}\right)$, we compute a token-level diff
\begin{equation}
    \Delta E_i^{(t)} = \textsc{Diff}\!\left(E_i^{(t)}, E_i^{(T)}\right),
\end{equation}
which identifies EC content that is deleted or replaced in the final version. Tokens in $E_i^{(t)}$ that correspond to deletions or substitutions in $\Delta E_i^{(t)}$ are treated as \emph{unstable spans}, representing eligibility criteria that were ultimately modified or removed.

Let $x_i^{(t)} = (x_{i,1}^{(t)}, \ldots, x_{i,T_i}^{(t)})$ denote the token sequence of $E_i^{(t)}$. During pretraining, CAMLM applies a span-aware masking strategy in which tokens within unstable spans are masked with a high probability $p_{\text{span}}$, while tokens outside these spans are masked with a lower background probability $p_{\text{low}}$. The model is trained using the standard MLM objective:
\begin{equation}
\mathcal{L}_{\text{CAMLM}} =
- \sum_{(i,t)} \sum_{m \in \mathcal{M}_i^{(t)}}
\log p\!\left(x_{i,m}^{(t)} \mid x_{i,\setminus m}^{(t)}\right),
\end{equation}
where $\mathcal{M}_i^{(t)}$ denotes the set of masked token positions for trial $i$ and version $t$. 
%This objective encourages the model to allocate greater representational capacity to amendment-relevant EC content while preserving general biomedical language understanding.

After CAMLM pretraining, the resulting encoder can be used either as a frozen feature extractor or fine-tuned end-to-end for eligibility-criteria amendment prediction as defined in Section~\ref{subsec:task_definition}. Because CAMLM introduces no additional parameters and makes no architectural changes, it can be applied directly to existing transformer encoder models like BERT, BioBERT, etc. In Section~\ref{subsec: prediction}, we show that CAMLM consistently improves amendment prediction performance across multiple model classes and datasets, with gains that persist under extensive ablation studies.

\begin{figure}[!htb]
    \centering
    \includegraphics[width=1.00\linewidth]{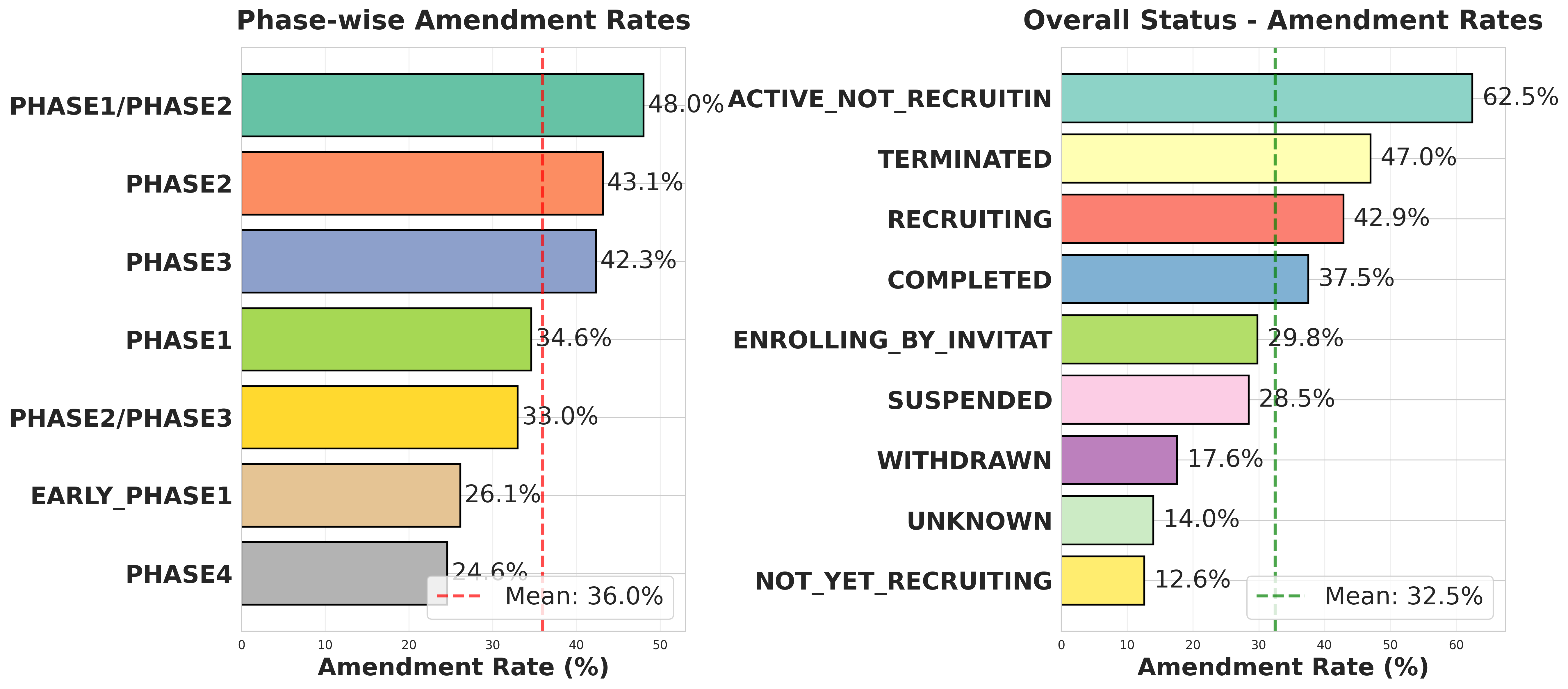}
    \caption{Clinical trial eligibility criteria amendment rates across phases and trial status in the \texttt{AMEND\_LLM} dataset (n=64,641) dataset. Dashed lines indicate mean amendment rates for each dimension.}
    \label{fig:amend_llm_stats}
\end{figure}

\begin{figure}[!htb]
    \centering
    \includegraphics[width=1.00\linewidth]{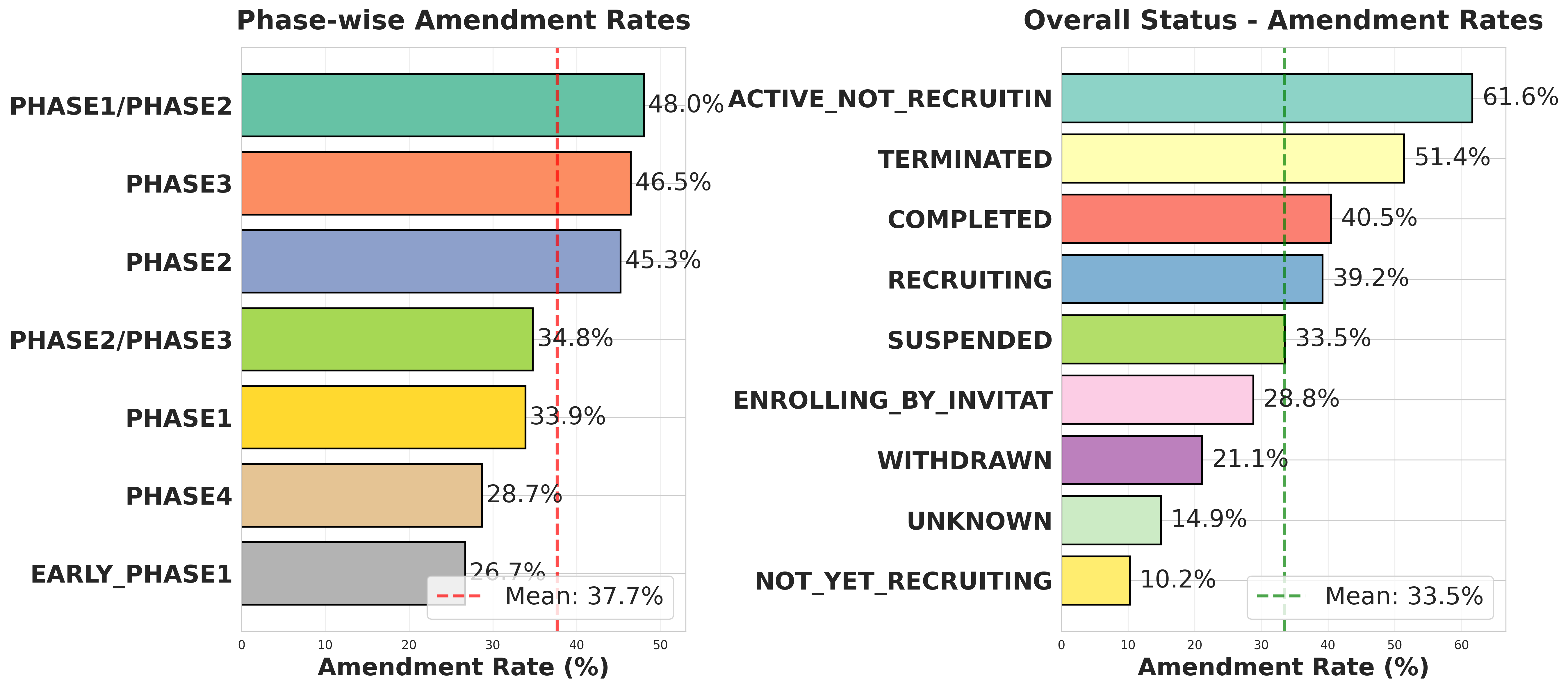}
    \caption{Clinical trial eligibility criteria amendment rates across phases and trial status in the \texttt{AMEND} (n=161,970) dataset. Dashed lines indicate mean amendment rates for each dimension.}
    \label{fig:amend_stats}
\end{figure}

\begin{table*}[t]
\centering
\caption{Performance comparison of different models on the \texttt{AMEND} dataset for eligibility criteria amendment prediction. Results are reported as mean $\pm$ standard deviation over multiple runs. \textbf{Bold} indicates the best score for each metric, and \underline{underlining} marks the better-performing model between CAMLM and its corresponding non-CAMLM variant.}

\label{tab:amend_model_comparison_amend}
\begin{tabular}{lccc}
\toprule
\textbf{Model} & AUROC & AUPRC & Accuracy \\
\midrule
BioBERT + LR 
& $0.676 \pm 0.007$ 
& $0.609 \pm 0.010$ 
& $0.629 \pm 0.006$ 
 \\
BioBERT\_CAMLM + LR 
& \underline{$0.689 \pm 0.007$} 
& \underline{$0.637\pm 0.010$} 
& \underline{$0.643 \pm 0.006$} 
 \\

\midrule
BioBERT + RF 
& $0.683 \pm 0.007$ 
& $0.616 \pm 0.011$ 
& $0.624 \pm 0.006$  \\

BioBERT\_CAMLM + RF 
& \underline{$0.684 \pm 0.007$} 
& \underline{$0.633\pm 0.010$} 
& \underline{$0.633 \pm 0.006$}  \\

\midrule
Finetuned BioBERT 
& $0.704 \pm 0.006$ 
& $0.656 \pm 0.009$ 
& $0.631 \pm 0.005$ \\

Finetuned BioBERT\_CAMLM 
& \underline{$\mathbf{0.714 \pm 0.007}$} 
& \underline{$\mathbf{0.670\pm 0.009}$} 
& \underline{$\mathbf{0.659 \pm 0.006}$} \\

\bottomrule
\end{tabular}
\end{table*}

\section{Experiments and Results}
\subsection{Summary of Datasets}
For this work, we focused exclusively on interventional drug trials. After filtering out trials with missing values, the \texttt{AMEND\_LLM} dataset contains 49,678 training, 8,896 validation, and 6,067 test trials, while the larger \texttt{AMEND} dataset includes 140,312 training, 15,591 validation, and the same 6,067 test trials. Both datasets share the test split to enable fair evaluation against human-labeled annotations. In addition to multiple versions of EC, both datasets provide rich trial-level metadata (i.e., disease area, intervention drugs, study phase, trial titles) that can serve as additional input features for amendment prediction models, helping to improve performance and contextual understanding. Detailed statistics of the datasets are shown in Table \ref{tab:dataset_stats}. Figure \ref{fig:amend_llm_stats} and Figure \ref{fig:amend_stats} show EC amendment rates across phases and trial status in \texttt{AMEND\_LLM} and \texttt{AMEND} datasets respectively.

\begin{table*}[t]
\centering
\caption{Performance comparison of different models on the \texttt{AMEND\_LLM} dataset for eligibility criteria amendment prediction. Results are reported as mean $\pm$ standard deviation over multiple runs. \textbf{Bold} indicates the best score for each metric, and \underline{underlining} marks the better-performing model between CAMLM and its corresponding non-CAMLM variant.}
\label{tab:amend_model_comparison_llm}
\begin{tabular}{lcccc}
\toprule
\textbf{Model} & AUROC & AUPRC & Accuracy \\
\midrule
BioBERT + LR 
& $0.681 \pm 0.007$ 
& $0.610 \pm 0.011$ 
& $0.633 \pm 0.006$ \\

BioBERT\_CAMLM + LR 
& \underline{$0.691 \pm 0.007$} 
& \underline{$0.632 \pm 0.010$} 
& \underline{$0.639 \pm 0.006$} \\

\midrule
BioBERT + RF 
& $0.676 \pm 0.007$ 
& $0.603 \pm 0.011$ 
& $0.617 \pm 0.006$ \\

BioBERT\_CAMLM + RF 
& \underline{$0.686 \pm 0.007$} 
& \underline{$0.628 \pm 0.010$} 
& \underline{$0.633 \pm 0.006$} \\

\midrule
Finetuned BioBERT
& $0.687 \pm 0.007$ 
& $0.621 \pm 0.010$ 
& $0.619 \pm 0.006$ \\

Finetuned BioBERT\_CAMLM
& \underline{$\mathbf{0.697 \pm 0.006}$}
& \underline{$\mathbf{0.634 \pm 0.010}$} 
& \underline{$\mathbf{0.644 \pm 0.006}$} \\

\bottomrule
\end{tabular}
\end{table*}

\begin{table*}[h!]
\centering
\caption{Ablations on pretraining strategies on \texttt{AMEND\_LLM}. Results shown for BioBERT + LR model.}
\label{tab:ablation}
\begin{tabular}{lccc}
\hline
Method & AUROC & AUPRC & Accuracy  \\
\hline
No pretraining & $0.681 \pm 0.007$ 
& $0.610 \pm 0.011$ 
& $0.633 \pm 0.006$\\
MLM & $0.687 \pm 0.007$ 
& $0.618 \pm 0.010$ 
& $0.635 \pm 0.006$ \\
Span MLM &  $0.688 \pm 0.007$ 
& $0.628 \pm 0.010$ 
& $0.636 \pm 0.006$ \\
CAMLM & $\mathbf{0.691 \pm 0.007}$ &  $\mathbf{0.632 \pm 0.010}$  
& $\mathbf{0.639 \pm 0.006}$ \\
\hline
\end{tabular}
\end{table*}

\subsection{Implementation Details}
For LLM-based denoising, we utilized the Qwen/Qwen2.5-7B-Instruct-1M model. All generations were performed with a temperature of 0.1 to ensure deterministic and stable outputs. The exact prompts used for denoising are provided in Appendix \ref{sec:appendix}. For extracting differences between two versions of EC, we utilized difflib python package \cite{python_difflib}. For benchmarking experiments, we employed BioBERT using the dmis-lab/biobert-base-cased-v1.1 checkpoint as the backbone encoder. We also used BERT and longformer for showing results using general and long context encoders respectively in the Appendix. All evaluation metrics are reported using nonparametric bootstrap resampling with 1,000 iterations over the test set, where we report the mean and standard deviation. Additional implementation details, including training configurations, hyperparameters and feature importance analysis are provided in Appendix.

\subsection{Eligibility Criteria Amendment Prediction} \label{subsec: prediction}

We evaluate \textit{eligibility criteria amendment prediction}, where models use the initial EC and other trial metadata available at trial initiation to predict whether EC will be amended later. 

As shown in Table~\ref{tab:amend_model_comparison_amend} and Table~\ref{tab:amend_model_comparison_llm}, frozen-embedding baselines achieve moderate performance on both datasets, indicating that static representations of initial eligibility criteria already capture some amendment-relevant signals. Among these baselines, BioBERT + Logistic Regression (LR) and BioBERT + Random Forest (RF) exhibit comparable performance when representations are fixed.

End-to-end fine-tuning with a BioBERT classifier consistently outperforms frozen-embedding baselines across both datasets, highlighting the importance of task-specific representation learning. While absolute performance is higher on the full \texttt{AMEND} dataset in most of the cases, incorporating CAMLM yields consistent gains across all model classes and metrics. For the fine-tuned models, CAMLM improves AUROC by approximately 1.4–1.5\%, AUPRC by about 2.1\%, and accuracy by 4.0–4.4\% relative on both datasets. \footnote{Relative improvements are computed as 
$(\text{CAMLM} - \text{baseline}) / \text{baseline} \times 100$.} Notably, these improvements are observed both on \texttt{AMEND} and \texttt{AMEND\_LLM}. 

\begin{comment}
This suggests that cleaner labels enable revision-aware pretraining to translate more reliably into downstream amendment prediction. At the same time, strong performance and consistent improvements on \texttt{AMEND} indicate that CAMLM remains robust under noisier, large-scale real-world supervision.     
\end{comment}

We evaluated the statistical significance of the AUROC differences using the DeLong's test. As shown in Table~\ref{tab:delong}, CAMLM pretraining yields statistically significant improvements over standard BioBERT for logistic regression and fine-tuned BioBERT models on both \texttt{AMEND} and \texttt{AMEND\_LLM} ($p<0.05$). No significant difference is observed for Random Forest on \texttt{AMEND}, while a significant gain is observed in \texttt{AMEND\_LLM}. As shown in Appendix Table~\ref{tab:amend_model_comparison_bert_longformer}, CAMLM pretraining consistently improves AUROC, AUPRC, and accuracy across different backbones (BERT, Longformer) and classifiers (LR, RF), demonstrating its generalizability beyond BioBERT. We focus on BioBERT-based baselines in the main paper because BioBERT is pretrained on biomedical text and consistently outperforms general-domain BERT and long-context Longformer models (see Appendix).

\subsection{Ablations}

To evaluate the impact of revision-aware pretraining, we compare four settings. (1) No pretraining: BioBERT is the off-the-shelf encoder. (2) MLM: further pretrains BioBERT with standard masked language modeling on the first EC version, capturing in-domain terminology but ignoring protocol evolution. (3) Span-aware MLM:  masks spans corresponding to textual changes between consecutive EC versions while lightly masking unchanged tokens, capturing short-range revision patterns. (4) CAMLM:  extends this to long-horizon modeling by pairing intermediate EC versions with the final protocol, focusing on regions predictive of eventual amendments. Table \ref{tab:ablation} shows that MLM on the first EC provides modest gains, span-aware MLM improves further by leveraging local changes, and CAMLM consistently outperforms all baselines, demonstrating that modeling EC evolution yields richer representations for amendment prediction than static text alone.

\subsection{Validity and Impact of LLM-Denoised Amendment Labels}

\begin{comment}
Given the high agreement between our LLM-generated labels and human annotations (Table \ref{tab:agreement}), and  the cost of large-scale human labeling, we directly use LLM-generated amendment labels for the training and validation splits in the \texttt{AMEND\_LLM} dataset. This allows us to scale reliable labels beyond what human annotation alone could provide. To ensure a high-quality evaluation benchmark, the test split was independently annotated by two clinical trial protocol experts (Algorithm \ref{alg:human_amendment}), with disagreements resolved via consensus discussion.    
\end{comment}

Given the substantially higher agreement between our LLM-generated labels and human annotations compared to both raw CTGov-derived labels and a baseline chain-of-thought (CoT) approach (Table~\ref{tab:agreement}), we directly use our LLM-generated amendment labels for the training and validation splits of the \texttt{AMEND\_LLM} dataset. Unlike the baseline CoT method, which treats amendment detection as a single-step prediction problem, our approach decomposes eligibility-criteria amendments into added, removed, and modified components, enabling more precise isolation of substantive eligibility changes. This structured decomposition yields markedly improved label fidelity while avoiding the prohibitive cost of large-scale human annotation, allowing us to scale reliable supervision beyond what manual labeling alone could support. To ensure a high-quality evaluation benchmark, the test split was independently annotated by two clinical trial protocol experts (Algorithm~\ref{alg:human_amendment}), with disagreements resolved through consensus discussion.

\begin{table}[!htb]
\centering
\caption{Agreement with Human-Annotated EC Amendment Labels (N=6067)}

\resizebox{\columnwidth}{!}{
\label{tab:agreement}
\begin{tabular}{lcc}
\hline
Method & Mismatches & Match Rate \\
\hline
CTGov Label & 230 & 96.21\% \\
Baseline CoT & 257 & 95.76\% \\
Our Approach & \textbf{40} & \textbf{99.34\%} \\
\hline
\end{tabular}}
\end{table}

\begin{table}[h!]
\centering
\small
\caption{Performance comparison of models trained on raw CTGov labels vs.\ LLM-denoised amendment labels, evaluated on human-labeled test data. These are BioBERT+LR results for showing the utility of the denoising step. Models are trained on \texttt{AMEND\_LLM} train split.}
\begin{tabular}{lcc}
\hline
Trained On & AUROC & AUPRC \\
\hline
CTGov labels 
& $0.672 \pm 0.007$ 
& $0.606 \pm 0.011$ \\

LLM labels 
& \textbf{0.681 $\pm$ 0.007} 
& \textbf{0.610 $\pm$ 0.011}\\
\hline
\end{tabular}
\label{tab:llm_denoising_vs_ctgov}
\end{table}

Table \ref{tab:llm_denoising_vs_ctgov} shows that replacing noisy CTGov\footnote{CTGov is short for \emph{ClinicalTrials.gov}.}
 labels with LLM-denoised labels consistently improves all evaluation metrics when tested against human annotations. The training and validation trials are from \texttt{AMEND\_LLM} as those have both CTGov labels and LLM-denoised labels. The test is the same test set annotated by human. This shows that denoising substantially enhances the quality of the label and yields better prediction performance.

\section{Conclusion}

We introduce \emph{eligibility criteria amendment prediction}, a novel NLP task addressing the frequent and impactful problem of clinical trial protocol amendments. To support this task, we release \texttt{AMEND++}, a benchmark suite comprising two datasets: \texttt{AMEND}, which captures eligibility-criteria version histories and amendment labels from public clinical trials, and \texttt{AMEND\_LLM}, a refined subset curated using an LLM-based denoising pipeline to isolate substantive changes. In addition, we propose CAMLM, a revision-aware pretraining strategy that takes advantage of historical edits to learn amendment-sensitive representations. Experiments demonstrate that CAMLM consistently improves prediction performance across datasets and architectures. Together, our datasets and pretraining approach offer a scalable framework for anticipating protocol amendments, enabling more efficient, cost-effective, and robust clinical trial design.

\section*{Limitations}
While we are the first to construct a machine-learning–ready dataset for protocol amendments, our work focuses exclusively on EC. EC is one of the most frequently amended and conceptually complex protocol sections, making it a natural starting point. However, other sections (such as outcomes, study design elements, and interventions) are also commonly amended, and extending our dataset and methods to these areas is an important direction for future work. A second limitation is our assumption that \textit{ClinicalTrials.gov} labels indicating no eligibility‐criteria amendment (label 0) are correct. We do not explicitly account for potential false negatives, which are expected to be rare in curated trial registries but could introduce residual label noise.

\bibliography{custom}

\begin{thebibliography}{17}
\providecommand{\natexlab}[1]{#1}

\bibitem[{Botto et~al.(2024)Botto, Smith, and Getz}]{botto2024new}
Emily Botto, Zachary Smith, and Kenneth Getz. 2024.
\newblock New benchmarks on protocol amendment experience in oncology clinical trials.
\newblock \emph{Therapeutic Innovation \& Regulatory Science}, 58(4):645--654.

\bibitem[{Chen et~al.(2024)Chen, Hu, Cai, Lu, Wang, Cao, Lin, Xu, Wu, Xiao et~al.}]{chen2024trialbench}
Jintai Chen, Yaojun Hu, Mingchen Cai, Yingzhou Lu, Yue Wang, Xu~Cao, Miao Lin, Hongxia Xu, Jian Wu, Cao Xiao, and 1 others. 2024.
\newblock Trialbench: Multi-modal artificial intelligence-ready clinical trial datasets.
\newblock \emph{arXiv preprint arXiv:2407.00631}.

\bibitem[{Cleophas et~al.(2009)Cleophas, Zwinderman, Cleophas, and Cleophas}]{cleophas2009statistics}
Ton~J Cleophas, Aeilko~H Zwinderman, Toine~F Cleophas, and Eugene~P Cleophas. 2009.
\newblock \emph{Statistics applied to clinical trials}.
\newblock Springer.

\bibitem[{Das et~al.(2025{\natexlab{a}})Das, Shafquat, Beigi, Aptekar, Mezey, and Sun}]{das2025seqtrial}
Trisha Das, Afrah Shafquat, Mandis Beigi, Jacob Aptekar, Jason Mezey, and Jimeng Sun. 2025{\natexlab{a}}.
\newblock Seqtrial: Utility preserving sequential clinical trial data generator.
\newblock In \emph{AMIA Annual Symposium Proceedings}, volume 2024, page 329.

\bibitem[{Das et~al.(2025{\natexlab{b}})Das, Shafquat, Beigi, Aptekar, and Sun}]{das-etal-2025-secret}
Trisha Das, Afrah Shafquat, Mandis Beigi, Jacob Aptekar, and Jimeng Sun. 2025{\natexlab{b}}.
\newblock \href {https://doi.org/10.18653/v1/2025.acl-long.264} {$secret$: Semi-supervised clinical trial document similarity search}.
\newblock In \emph{Proceedings of the 63rd Annual Meeting of the Association for Computational Linguistics (Volume 1: Long Papers)}, pages 5278--5291, Vienna, Austria. Association for Computational Linguistics.

\bibitem[{Das et~al.(2023)Das, Wang, and Sun}]{das2023twin}
Trisha Das, Zifeng Wang, and Jimeng Sun. 2023.
\newblock Twin: Personalized clinical trial digital twin generation.
\newblock In \emph{Proceedings of the 29th ACM SIGKDD Conference on Knowledge Discovery and Data Mining}, pages 402--413.

\bibitem[{Foundation(2025)}]{python_difflib}
Python~Software Foundation. 2025.
\newblock The `difflib` module — `difflib` — python 3.12.0 documentation.
\newblock \url{https://docs.python.org/3/library/difflib.html}.
\newblock Accessed: 2025-11-25.

\bibitem[{Fu et~al.(2022)Fu, Huang, Xiao, Glass, and Sun}]{fu2022hint}
Tianfan Fu, Kexin Huang, Cao Xiao, Lucas~M Glass, and Jimeng Sun. 2022.
\newblock Hint: Hierarchical interaction network for clinical-trial-outcome predictions.
\newblock \emph{Patterns}, 3(4).

\bibitem[{Gao et~al.(2025)Gao, Pradeepkumar, Das, Thati, and Sun}]{gao2025auto}
Chufan Gao, Jathurshan Pradeepkumar, Trisha Das, Shivashankar Thati, and Jimeng Sun. 2025.
\newblock \href {https://arxiv.org/abs/2406.10292} {Automatically labeling clinical trial outcomes: A large-scale benchmark for drug development}.
\newblock \emph{Preprint}, arXiv:2406.10292.

\bibitem[{Getz et~al.(2024)Getz, Smith, Botto, Murphy, and Dauchy}]{getz2024new}
Kenneth Getz, Zachary Smith, Emily Botto, Elisabeth Murphy, and Arnaud Dauchy. 2024.
\newblock New benchmarks on protocol amendment practices, trends and their impact on clinical trial performance.
\newblock \emph{Therapeutic Innovation \& Regulatory Science}, 58(3):539--548.

\bibitem[{Getz et~al.(2016)Getz, Stergiopoulos, Short, Surgeon, Krauss, Pretorius, Desmond, and Dunn}]{getz2016impact}
Kenneth~A Getz, Stella Stergiopoulos, Mary Short, Leon Surgeon, Randy Krauss, Sybrand Pretorius, Julian Desmond, and Derek Dunn. 2016.
\newblock The impact of protocol amendments on clinical trial performance and cost.
\newblock \emph{Therapeutic innovation \& regulatory science}, 50(4):436--441.

\bibitem[{Joshi(2023)}]{joshi2023common}
Shivam Joshi. 2023.
\newblock Common clinical trial amendments, why they are submitted and how they can be avoided: a mixed methods study on nhs uk sponsored research (amendments assemble).
\newblock \emph{Trials}, 24(1):10.

\bibitem[{Lewis(1999)}]{Lewis1999StatisticalPFA}
John~A. Lewis. 1999.
\newblock \href {https://www.ncbi.nlm.nih.gov/pubmed/10440877} {Statistical principles for clinical trials (ich e9): an introductory note on an international guideline.}
\newblock \emph{Statistics in medicine}, 18 15:1903--42.

\bibitem[{L{\"o}sch and Neuh{\"a}user(2008)}]{losch2008statistical}
Christian L{\"o}sch and Markus Neuh{\"a}user. 2008.
\newblock The statistical analysis of a clinical trial when a protocol amendment changed the inclusion criteria.
\newblock \emph{BMC medical research methodology}, 8(1):16.

\bibitem[{Luo et~al.(2024)Luo, Qian, Glass, and Ma}]{luo2024clinical}
Junyu Luo, Cheng Qian, Lucas Glass, and Fenglong Ma. 2024.
\newblock Clinical trial retrieval via multi-grained similarity learning.
\newblock In \emph{Proceedings of the 47th International ACM SIGIR Conference on Research and Development in Information Retrieval}, pages 2950--2954.

\bibitem[{Tseo et~al.(2020)Tseo, Salkola, Mohamed, Kumar, and Abnousi}]{tseo2020information}
Yitong Tseo, MI~Salkola, Ahmed Mohamed, Anuj Kumar, and Freddy Abnousi. 2020.
\newblock Information extraction of clinical trial eligibility criteria.
\newblock \emph{arXiv preprint arXiv:2006.07296}.

\bibitem[{Yuan et~al.(2019)Yuan, Ryan, Ta, Guo, Li, Hardin, Makadia, Jin, Shang, Kang et~al.}]{yuan2019criteria2query}
Chi Yuan, Patrick~B Ryan, Casey Ta, Yixuan Guo, Ziran Li, Jill Hardin, Rupa Makadia, Peng Jin, Ning Shang, Tian Kang, and 1 others. 2019.
\newblock Criteria2query: a natural language interface to clinical databases for cohort definition.
\newblock \emph{Journal of the American Medical Informatics Association}, 26(4):294--305.

\end{thebibliography}
\newpage
\appendix
\section*{Appendix A}
\label{sec:appendix}

\subsection*{Hyperparameter Tuning}
For the BioBERT + LR baseline, we tuned C (inverse of the regularization strength) from [0.01, 0.1, 1, 10, 100]. For the BioBERT + RF baseline, we tuned n\_estimators from [100, 200, 300] and max\_depth from [5, 10]. We did not let any weight updates of BioBERT for these baselines. For finetuning the BioBERT model, we used 10 epochs for training. We selected a model based on the validation AUROC evaluated after each epoch. 

For CAMLM, MLM and Span MLM, we trained BioBERT for 3 epochs. For CAMLM pretraining, we use BioBERT’s WordPiece tokenizer. Change spans are computed at the token level using \texttt{difflib.SequenceMatcher} between the tokenized initial and final EC versions, where \texttt{replace} and \texttt{delete} operations define unstable spans. We use a \emph{change-aware} masked language modeling (MLM) objective that emphasizes amendment-prone regions, masking tokens within detected change spans with probability $p_{\text{span}} = 0.8$ and masking remaining tokens with a lower background probability $p_{\text{low}} = 0.05$. All masked tokens are replaced with the standard \texttt{[MASK]} token, and supervision is provided only at masked positions following standard MLM training.

\begin{table*}[!h]
\centering
\caption{Feature Importance Exploration}
\label{tab:feature_importance_biobert_lr}
\resizebox{\textwidth}{!}{
\begin{tabular}{lcccc}
\hline
\textbf{Features Used} & AUROC & AUPRC & Accuracy & F1 \\
\hline
First\_EC & 0.675 $\pm$ 0.007 & 0.607 $\pm$ 0.011 & 0.630 $\pm$ 0.006 & 0.563 $\pm$ 0.008 \\
First\_EC, disease & 0.676 $\pm$ 0.007 & 0.609 $\pm$ 0.011 & 0.630 $\pm$ 0.006 & 0.563 $\pm$ 0.008 \\
First\_EC, disease, intervention & 0.680 $\pm$ 0.007 & 0.606 $\pm$ 0.011 & 0.633 $\pm$ 0.006 & 0.567 $\pm$ 0.008 \\
First\_EC, disease, intervention, phase & 0.681 $\pm$ 0.007 & 0.610 $\pm$ 0.011 & 0.633 $\pm$ 0.006 & 0.567 $\pm$ 0.008 \\
\hline
\end{tabular}}
\end{table*}

\subsection*{Feature Importance}
We tried to understand feature importance on the BioBERT + LR model. We assume that the findings from the results will be similar for other models too. Table \ref{tab:feature_importance_biobert_lr} shows that using the first version of EC, disease/condition, intervention and phase combined gives the highest scores.

\begin{algorithm*}[!ht]
\caption{Human Annotation Procedure for Determining EC Amendments}
\label{alg:human_amendment}
\DontPrintSemicolon

\KwIn{First version of eligibility criteria $EC_{\text{first}}$,\\
\hspace{1.8em} Final version of eligibility criteria $EC_{\text{final}}$,\\
\hspace{1.8em} \texttt{ctgov\_label} $\in \{0,1\}$}

\KwOut{\texttt{amendment} $\in \{0,1\}$}

\BlankLine

\If{\texttt{ctgov\_label} $= 0$}{
    \Return 0 \tcp*[r]{Use ctgov label as ground truth}
}

\SetKw{Continue}{continue}
\SetKw{Break}{break}

\BlankLine
\textbf{Initialize:} $\texttt{amendment} \gets 0$\;

\BlankLine
\textbf{Rule 1: Addition Check}\;
\If{there exists a criterion in $EC_{\text{final}}$ not present in $EC_{\text{first}}$}{
    $\texttt{amendment} \gets 1$\;
}

\BlankLine
\textbf{Rule 2: Removal Check}\;
\If{there exists a criterion in $EC_{\text{first}}$ not present in $EC_{\text{final}}$}{
    $\texttt{amendment} \gets 1$\;
}

\BlankLine
\textbf{Rule 3: Significant Modification Check}\;
\If{any criterion is modified such that:}{
    numeric thresholds change \textbf{or} medical entities are added/removed \textbf{or}\;
    the meaning of the criterion changes or the affected population differs\;
}{
    $\texttt{amendment} \gets 1$\;
}

\BlankLine
\textbf{Ignored non-amendments:}\;
\Indp
Formatting differences (bullets, numbering, indentation)\;
Minor spelling or grammar corrections\;
Synonyms or wording changes with preserved meaning\;
Abbreviation expansion or contraction (e.g., ECG $\rightarrow$ electrocardiogram)\;
Reordering of criteria without semantic change\;
Text added or removed that does not affect eligibility or population\; etc\;
\Indm

\BlankLine
\Return $\texttt{amendment}$\;

\end{algorithm*}

\begin{table*}[t]
\centering
\small
\begin{tabular}{lcccc}
\toprule
\multirow{2}{*}{Comparison} 
& \multicolumn{2}{c}{\textbf{AMEND}} 
& \multicolumn{2}{c}{\textbf{AMEND\_LLM}} \\
\cmidrule(lr){2-3} \cmidrule(lr){4-5}
& p-value & Sig. & p-value & Sig. \\
\midrule
BioBERT (LR) vs. BioBERT-CAMLM (LR)
& $3.24\times10^{-3}$ & \checkmark 
& $3.03\times10^{-2}$ & \checkmark \\

BioBERT (RF) vs. BioBERT-CAMLM (RF)
& $9.24\times10^{-1}$ & -- 
& $4.87\times10^{-3}$ & \checkmark \\

Finetuned BioBERT vs. Finetuned BioBERT-CAMLM
& $1.23\times10^{-3}$ & \checkmark 
& $5.05\times10^{-3}$ & \checkmark \\
\bottomrule
\end{tabular}
\caption{Statistical significance of AUROC differences between BioBERT and BioBERT-CAMLM variants, evaluated using DeLong’s test.}
\label{tab:delong}
\end{table*}

\begin{table*}[t]
\centering
\caption{Performance comparison of all baseline models with and without CAMLM on the \texttt{AMEND\_LLM} dataset. Results are reported as mean $\pm$ standard deviation over multiple runs. \textbf{Bold} indicates the best score for each metric, and \underline{underlining} marks the better-performing model between CAMLM and its corresponding non-CAMLM variant.}
\label{tab:amend_model_comparison_bert_longformer}
\begin{tabular}{lccc}
\toprule
\textbf{Model} & \textbf{AUROC} & \textbf{AUPRC} & \textbf{Accuracy} \\
\midrule
BERT + LR 
& $0.669 \pm 0.007$ 
& $0.602 \pm 0.010$ 
& $0.616 \pm 0.006$ \\

BERT\_CAMLM + LR 
& \underline{$0.675 \pm 0.007$} 
& \underline{$0.620 \pm 0.010$} 
& \underline{$0.624 \pm 0.006$} \\

\midrule
BERT + RF 
& $0.660 \pm 0.007$ 
& $0.598 \pm 0.010$ 
& $0.605 \pm 0.006$ \\

BERT\_CAMLM + RF 
& \underline{$0.674 \pm 0.007$} 
& \underline{$0.616 \pm 0.010$} 
& \underline{$0.618 \pm 0.006$} \\

\midrule
Longformer + LR 
& $0.670 \pm 0.007$ 
& $0.611 \pm 0.010$ 
& $0.622 \pm 0.006$ \\

Longformer\_CAMLM + LR 
& \underline{$0.674 \pm 0.007$} 
& \underline{$0.618 \pm 0.010$} 
& \underline{$0.626 \pm 0.006$} \\

\midrule
Longformer + RF 
& $0.666 \pm 0.007$ 
& $0.602 \pm 0.010$ 
& $0.597 \pm 0.006$ \\

Longformer\_CAMLM + RF 
& \underline{$0.673 \pm 0.007$} 
& \underline{$0.627 \pm 0.010$} 
& \underline{$0.619 \pm 0.006$} \\
\midrule

BioBERT + LR 
& $0.681 \pm 0.007$ 
& $0.610 \pm 0.011$ 
& $0.633 \pm 0.006$ \\

BioBERT\_CAMLM + LR 
& \underline{$0.691 \pm 0.007$} 
& \underline{$0.632 \pm 0.010$} 
& \underline{$0.639 \pm 0.006$} \\

\midrule
BioBERT + RF 
& $0.676 \pm 0.007$ 
& $0.603 \pm 0.011$ 
& $0.617 \pm 0.006$ \\

BioBERT\_CAMLM + RF 
& \underline{$0.686 \pm 0.007$} 
& \underline{$0.628 \pm 0.010$} 
& \underline{$0.633 \pm 0.006$} \\

\midrule
Finetuned BioBERT
& $0.687 \pm 0.007$ 
& $0.621 \pm 0.010$ 
& $0.619 \pm 0.006$ \\

Finetuned BioBERT\_CAMLM
& \underline{$\mathbf{0.697 \pm 0.006}$}
& \underline{$\mathbf{0.634 \pm 0.010}$} 
& \underline{$\mathbf{0.644 \pm 0.006}$} \\

\bottomrule
\end{tabular}
\end{table*}

\subsection*{LLM Prompts}
Figure \ref{fig: criteria_added}, figure \ref{fig: criteria_modified} and figure \ref{fig: criteria_removed} are prompts used for denoising amendment labels.

\subsection*{Usage of AI Assistants}
We utilized chatgpt for paper drafting (e.g., summarizing, paraphrasing, etc.) assistance.

\begin{figure}[htbp]
\centering
\begin{tcolorbox}[
    colback=gray!10,
    colframe=gray!80,
    boxrule=0.5pt,
    arc=3pt,
    left=6pt, right=6pt, top=6pt, bottom=6pt,
    title={Prompt for "Criteria Removed" Task},
    fonttitle=\bfseries,
    fontupper=\small\ttfamily,
    halign=flush left,
    breakable
]
Below is an example diff between two versions of eligibility criteria (EC), with correct reasoning.

\vspace{1em}
\{example\}
\vspace{1em}

Now apply the same reasoning to the new case below.

Here are the differences between two versions of eligibility criteria:

\vspace{1em}
\{difference\}
\vspace{1em}

Your thoughts:

You are a clinical trial analyst. Think step-by-step about the differences between two versions of eligibility criteria (EC) and determine if any existing criteria have been removed.

\vspace{1em}
Step-by-step Instructions:

\begin{itemize}[label=-, leftmargin=*, noitemsep, topsep=2pt]
    \item Read all lines starting with '-' as a single block. These represent the original eligibility criteria.
    \item Read all lines starting with '+' as a single block. These represent the updated eligibility criteria.
    \item Compare the meaning of the '-' block with the '+' block as a whole.
    \item Ignore formatting changes (e.g., line breaks, indentation) and rewordings that preserve the original meaning.
    \item Focus \textbf{only} on whether the '-' block contains any criteria that are completely missing in the '+' block.
    \item If any requirement, rule, or constraint was present in the '-' block but is no longer found in the '+' block, then a criterion has been removed.
\end{itemize}

\vspace{1em}
If at least one criterion has been removed:\\
Final output: 1

If no criteria have been removed:\\
Final output: 0

Think step-by-step.

Your final answer must be a single line at the end in the following format (case-sensitive):

Final output: 0\\
or\\
Final output: 1

\end{tcolorbox}
\caption{Prompt given to the LLM for the ``Criteria Removed'' task.}
\label{fig: criteria_removed}
\end{figure}

\begin{figure*}[htbp]
\centering
\begin{tcolorbox}[
    colback=gray!10,
    colframe=gray!80,
    boxrule=0.5pt,
    arc=3pt,
    left=6pt, right=6pt, top=6pt, bottom=6pt,
    title={Prompt for "Criteria Modified" Task},
    fonttitle=\bfseries,
    fontupper=\small\ttfamily,
    halign=flush left,
    breakable
]
Below is an example diff between two versions of eligibility criteria (EC), with correct reasoning.

\vspace{1em}
\{example\}
\vspace{1em}

Now apply the same reasoning to the new case below.

Here are the differences between two versions of eligibility criteria:

\vspace{1em}
\{difference\}
\vspace{1em}

Your thoughts:

You are a clinical trial analyst. Your task is to detect whether any existing eligibility criteria have been *modified* in a way that changes who is eligible.
You are not responsible for detecting new criteria or removed criteria - those are handled separately.

\vspace{1em}
Instructions:

\begin{itemize}[label=-, leftmargin=*, noitemsep, topsep=2pt]
    \item Treat '-' lines as the original version and '+' lines as the updated version of the *same* criterion.
    \item Compare carefully for semantic differences.
\end{itemize}

\vspace{1em}
Count as meaningful changes (Final output: 1):
\begin{itemize}[label=-, leftmargin=*, noitemsep, topsep=2pt]
    \item Numeric changes (age limits, lab thresholds, dosage values).
    \item Severity qualifiers (e.g., "liver disease" --> "severe liver disease").
    \item Logical operator changes (e.g., "and" <--> "or").
    \item Added or removed biomedical entities within the same criterion:
    \begin{itemize}[label=*, leftmargin=1.5em, noitemsep]
        \item Drugs (e.g., "warfarin" --> "warfarin or heparin").
        \item Diseases/conditions (e.g., "HIV" --> "HIV or HBV").
        \item Enzymes/transporters/biomarkers (e.g., "CYP3A4" --> "CYP3A4 or P-gp").
    \end{itemize}
    \item Time window changes (e.g., "within 6 months" --> "within 3 months").
    \item Condition scope changes (e.g., "cancer" --> "breast cancer", "diabetes" --> "type 2 diabetes").
    \item Entity substitutions (e.g., "rituximab" --> "adalimumab").
\end{itemize}

Ignore as non-meaningful (Final output: 0):
\begin{itemize}[label=-, leftmargin=*, noitemsep, topsep=2pt]
    \item Rearrangement of criteria or bullet points.
    \item Capitalization differences (e.g., "BOTH" --> "both").
    \item Formatting or indentation changes.
    \item Punctuation only (commas, semicolons, periods).
    \item Abbreviation expansion or contraction (e.g., "ART" <--> "antiretroviral therapy (ART)", "CYP3A4" <--> "cytochrome P450 3A4").
    \item Synonyms with no scope change (e.g., "high blood pressure" <--> "hypertension").
    \item Spelling corrections or typos (e.g., "diabtes" --> "diabetes").
    \item Minor stylistic rephrasing without meaning change (e.g., "patients who have" --> "patients with").
    \item Splitting or merging sentences without altering meaning.
    \item Reordering of biomedical terms without additions/removals (e.g., "HIV, HBV, HCV" <--> "HBV, HCV, HIV").
\end{itemize}

\vspace{1em}
Output format (case sensitive):
\begin{itemize}[label=-, leftmargin=*, noitemsep, topsep=2pt]
    \item If at least one meaningful modification is found:\\
    Final output: 1
    \item If no meaningful modification is found:\\
    Final output: 0
\end{itemize}

\end{tcolorbox}
\caption{Prompt given to the LLM for the ``Criteria Modified'' task.}
\label{fig: criteria_modified}
\end{figure*}

%-------------------
% Two-column wide prompt box for "criteria added"
\begin{figure*}[htbp]
\centering
\begin{tcolorbox}[
    colback=gray!10,
    colframe=gray!80,
    boxrule=0.5pt,
    arc=3pt,
    left=6pt, right=6pt, top=6pt, bottom=6pt,
    title={Prompt for "Criteria Added" Task},
    fonttitle=\bfseries,       % Makes the title bold
    fontupper=\small\ttfamily, % Sets the body font to small monospace
    halign=flush left,         % Aligns text left (better for mono fonts)
    breakable                  % Allows the box to split across pages if needed
]

Below is an example diff between two versions of eligibility criteria (EC), with correct reasoning.

\vspace{1em}
\{example\}
\vspace{1em}

Now apply the same reasoning to the new case below.
Here are the differences between two versions of eligibility criteria:

\vspace{1em}
\{difference\}
\vspace{1em}

Your thoughts:

You are a clinical trial analyst. Think step-by-step about the differences between two versions of eligibility criteria (EC) and determine if any new criteria have been added.

\vspace{1em}
Step-by-step Instructions:

% Using itemize makes the wrapping look much cleaner
\begin{itemize}[label=-, leftmargin=*, noitemsep, topsep=2pt]
    \item Read all lines starting with '-' as a single block. These represent the original eligibility criteria.
    \item Read all lines starting with '+' as a single block. These represent the updated eligibility criteria.
    \item Compare the meaning of the '+' block with the '-' block as a whole.
    \item Ignore formatting changes (e.g., line breaks, indentation) and rewordings that preserve the original meaning.
    \item Focus only on whether the '+' block introduces any new criterion that is not already present in the '-' block.
    \item If any new requirement, rule, or constraint appears in the '+' block that is absent from the '-' block, it is a new criterion and constitutes an amendment.
\end{itemize}

\vspace{1em}
If at least one new criterion is added:\\
Final output: 1

If no new criteria are added:\\
Final output: 0

Think step-by-step.

Your final answer must be a single line at the end in the following format (case-sensitive):

Final output: 0\\
or\\
Final output: 1

\end{tcolorbox}
\caption{Prompt given to the LLM for the ``Criteria Added'' task.}
\label{fig: criteria_added}
\end{figure*}

\end{document}